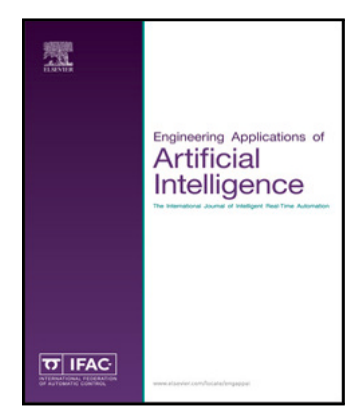

Research paper

# Adaptive Kalman-Informed Transformer

Nadav Cohen *, Itzik Klein

*The Hatter Department of Marine Technologies, Charney School of Marine Sciences, University of Haifa, Haifa, Israel*

ABSTRACT

The extended Kalman filter (EKF) is a widely adopted method for sensor fusion in navigation applications. A crucial aspect of the EKF is the online determination of the process noise covariance matrix reflecting the model uncertainty. While common EKF implementation assumes a constant process noise, in real-world scenarios, the process noise varies, leading to inaccuracies in the estimated state and potentially causing the filter to diverge. Model-based adaptive EKF methods were proposed and demonstrated performance improvements to cope with such situations, highlighting the need for a robust adaptive approach. In this paper, we derive an adaptive Kalman-informed transformer (A-KIT) designed to learn the varying process noise covariance online. Built upon the foundations of the EKF, A-KIT utilizes the well-known capabilities of set transformers, including inherent noise reduction and the ability to capture nonlinear behavior in the data. This approach is suitable for any application involving the EKF. In a case study, we demonstrate the effectiveness of A-KIT in nonlinear fusion between a Doppler velocity log and inertial sensors. This is accomplished using real data recorded from sensors mounted on an autonomous underwater vehicle operating in the Mediterranean Sea. We show that A-KIT outperforms the conventional EKF by more than 49.5% and model-based adaptive EKF by an average of 35.4% in terms of position accuracy.

## 1. Introduction

The Kalman filter (KF) is renowned for its effectiveness in data fusion. It is acknowledged as an optimal state estimator designed for linear dynamic systems affected by Gaussian white noise (Bar-Shalom et al., 2001). In real-life applications, the majority of systems exhibit nonlinear behavior. Consequently, various extensions of the KF have been developed, such as the extended Kalman filter (EKF) (Ribeiro, 2004), unscented Kalman filter (UKF) (Wan and Van Der Merwe, 2000), and others (Daum, 2005). These extensions are designed to effectively address nonlinear systems, albeit at the cost of sub-optimality. As a result, variations of the KF have found applications in diverse fields, including tracking, navigation, control, signal processing, and communication.

In the conventional KF framework, it is assumed that both process and measurement noise covariance matrices are known, well-defined and play a crucial role in conjunction with system dynamics to dictate the achievable accuracy of the filter (Heffes, 1966; Sangsuk-Iam and Bullock, 1990; Saab, 1995). However, in numerous practical scenarios, particularly when dealing with noisy feature data in nonlinear systems, the statistical properties of the noise covariances are frequently unknown or only partially discernible. Moreover, these properties may undergo changes throughout the duration of a mission (Simon, 2006; Farrell, 2008). As a result, noise identification becomes a critical component, giving rise to adaptive filtering.

The exploration of model-based adaptive KF dates back to 1970, with Mehra (Mehra, 1970) introducing an approach to address optimal filtering for a linear time-invariant system characterized by unknown process noise covariance. Subsequently, additional approaches emerged, categorized into four distinct groups: Bayesian inference, maximum likelihood estimation, covariance-matching, and correlation methods. These approaches enhance state estimation performance in environments where noise characteristics are not constant, providing more accurate and reliable estimates compared to traditional Kalman filters. (Zhang et al., 2020).

Recent advances in hardware and computational efficiency have demonstrated the utility of deep learning (DL) methods in addressing real-time applications. These applications span various domains, including image processing, signal processing, and natural language processing. Leveraging the inherent capabilities of DL to handle nonlinear problems, these methods have found integration into inertial navigation algorithms. Accordingly, research initiatives have been initiated to investigate the data-driven estimation of the process noise covariance matrix across various domains, including land, aerial, and maritime applications (Li et al., 2021; Cohen and Klein, 2024).

* Corresponding author.
*E-mail address:* ncohe140@campus.haifa.ac.il (N. Cohen).

In the standard navigation setting, a key method for obtaining an accurate and reliable navigation solution involves the integration of measurements from various sensors through the use of a nonlinear filter, such as the EKF. The inertial measurement unit (IMU) data is processed through the inertial navigation system (INS) equations of motion to derive a continuous, high-rate navigation solution. This system, however, is characterized by error accumulation due to the inherent nature of the inertial sensor. To mitigate this error, the EKF relies on external observations from other sensors, such as the global navigation satellite system (GNSS) or Doppler velocity log (DVL), to enhance the accuracy of the navigation solution (Chatfield, 1997; Titterton et al., 2004; Abosekeen et al., 2019).

In this paper, we propose a novel approach named A-KIT, an adaptive Kalman-informed transformer. Built upon a set-transformer network, A-KIT is designed for real-time adaptive regression of the process noise covariance matrix. Using a set-transformer for estimating process noise covariance offers advantages such as permutation invariance, ensuring consistent results when handling asynchronously arriving data or data without a meaningful sequence. It effectively models complex dependencies by capturing correlations between system states or sensors through self-attention. Furthermore, it demonstrates resilience to noise and missing data by emphasizing the most relevant inputs using attention mechanisms. The set-transformer generalizes well, integrates heterogeneous data sources, and avoids strong distributional assumptions, making it a promising choice for learning covariance in complex, high-dimensional, or nonlinear systems as will be discussed in more detail throughout the paper. The study showcases A-KIT's superiority, outperforming both the conventional EKF and various state-of-the-art, model-based, adaptive EKF approaches in an INS/DVL fusion case study. The key contributions of this paper include:

1. The introduction of A-KIT, a hybrid algorithm combining the strengths of the well-established EKF and leveraging well-known DL characteristics, such as noise reduction, for improved performance.
2. A novel Kalman-informed loss configuration designed to emulate the principles of the KF, enhancing the accuracy of the process noise covariance.
3. A GitHub repository containing experiment data and code for implementing the A-KIT architecture. The repository serves as a comprehensive resource for researchers and practitioners interested in replicating or further exploring the proposed A-KIT algorithm.

The remainder of the paper is organized as follows: In Section 2, we discuss related work. Section 3 delves into the theoretical and mathematical background of model-based adaptive estimation. Section 4 introduces the components of the set-transformer architecture. In Section 5, we present the proposed A-KIT approach. Section 6 establishes the case study of the INS/DVL sensor fusion and discusses the implementation and integration of A-KIT into this problem. In Section 7, the dataset acquisition process is detailed, along with an in-depth analysis of the results. Finally, Section 8 discusses the conclusions.

## 2. Related work

When examining model-based adaptive KF approaches, various studies have demonstrated improvements in INS and GNSS navigation across different platforms using innovation-based adaptive KF (Mohamed and Schwarz, 1999; Hide et al., 2003; Liu et al., 2018; Shen et al., 2022). Additionally, the effectiveness of adaptive KFs in INS/DVL applications has been showcased in works such as (Gao et al., 2015; Wang et al., 2020; Silva et al., 2022; Wang et al., 2023; Or and Klein, 2023b).

For land-based navigation in the context of terrestrial vehicles, Brossard, Barrau, and Bonnabel proposed a method that utilizes convolutional neural networks (CNNs) to dynamically adjust the covariance noise matrix within an invariant EKF. This adaptation is achieved by incorporating moderately priced IMU measurements (Brossard et al., 2020). In an alternative approach, specifically designed for land vehicles, reinforcement learning takes center stage, as elucidated in Gao et al. (2020). Wu et al. (2020) conducted a study where not only were the parameters of measurement noise covariances regressed, but also the parameters of process noise covariances were predicted. This prediction utilized a multitask temporal convolutional network (TCN), resulting in superior position accuracy compared to traditional GNSS/INS integrated navigation systems. In a unique approach, Xiao et al. (2021) introduced a residual network incorporating an attention mechanism specifically designed to predict individual velocity elements of the noise covariance matrix, with a focus on land vehicle applications.

For aerial platforms, a CNN-based adaptive Kalman filter was proposed by Zou et al. (2020) to enhance high-speed navigation using low-cost IMU. The study introduces a 1D CNN approach to predict 3D acceleration and angular velocity noise covariance information. In subsequent work, Or and Klein Or and Klein (2022a) introduced a data-driven, adaptive noise covariance approach for an error state EKF in INS/GNSS fusion. The information obtained was utilized to enhance the navigation of a quadrotor drone. In the visual-inertial odometry (VIO) task for micro aerial vehicles, dynamically estimating the inertial noise uncertainty using a dedicated deep neural network improved the accuracy and robustness of the VIO algorithm (Solodar and Klein, 2024).

When examining autonomous underwater vehicle (AUV) navigation with INS/DVL sensor fusion, Huang et al. (2023) proposed the utilization of a long short-term memory (LSTM) network to estimate the initial and constant process noise covariance matrix. Their work demonstrated significant improvement over conventional EKF approaches. Or and Klein (2022a,b, 2023b) devised an adaptive EKF customized for velocity updates in INS/DVL fusions. Initially, they highlighted the efficacy of correcting the noise covariance matrix through a 1D CNN, utilizing a classification approach to predict the variance at each sample time. This correction method resulted in substantial improvements in navigation results. In their most recent work, the authors introduced a novel approach that employs regression instead of classification to achieve the same objective. However, a crucial observation must be noted, while the approach demonstrated success when tested on simulated data, its performance did not translate effectively when applied to real-world data obtained from an AUV.

In the exploration of related work, Haarnoja et al. (2016) were pioneers in examining a closed-loop KF solution. They approached the problem by drawing parallels to recurrent neural networks (RNNs) and highlighted the necessity of implementing a back-propagation through time approach, emphasizing its computational cost. Their methodology was tested for solving the visual odometry problem, and the designed 1D CNN provided the updated measurement and the measurement noise covariance. In Jouaber et al. (2021), the authors address a constant velocity model in a linear Kalman Filter for tracking applications. This approach is evaluated solely on simulated data. Additional work on process noise covariance estimation in a linear Kalman filter addressed the problem of vehicle trajectory uncertainty (Or and Klein, 2023a).

## 3. Model-based adaptive estimation

A prevalent method for conducting state estimation with a given model and set of observations is the KF. In accordance with the theory, the filter leads to the optimal linear state estimator that minimizes the mean square error under Gaussian assumptions (Bar-Shalom et al., 2001; Simon, 2006). However, assuming a linear model in real-world applications such as navigation and control is unrealistic, as these scenarios typically involve nonlinear problems. To address this, an extension of the linear Kalman Filter, known as the EKF, capable of handling nonlinear problems, was introduced (Groves, 2015). In this section, a concise theoretical background is presented, introducing both the error-state EKF and the adaptive error-state EKF.

### 3.1. Conventional Extended Kalman Filter

We describe the error-state implementation of the EKF (ES-EKF). To that end, consider $\delta\boldsymbol{x} \in \mathbb{R}^{n\times 1}$, an $n$-dimensional error-state vector, defined as follows:

$$\boldsymbol{x}^t = \boldsymbol{x}^e - \delta\boldsymbol{x} \tag{1}$$

where $\boldsymbol{x}^t \in \mathbb{R}^{n\times 1}$ and $\boldsymbol{x}^e \in \mathbb{R}^{n\times 1}$ are the true state and the estimated state, respectively. In the EKF, the underlying assumption is that the error in the state vector estimate is substantially smaller than the state vector. This condition facilitates the application of a linear system model to the state vector residual:

$$\delta\dot{\boldsymbol{x}} = \mathbf{F}\delta\boldsymbol{x} + \mathbf{G}\boldsymbol{n} \tag{2}$$

where $\boldsymbol{n} \in \mathbb{R}^{n\times 1}$ is the system noise vector, $\mathbf{F} \in \mathbb{R}^{n\times n}$ is the system matrix, and $\mathbf{G} \in \mathbb{R}^{n\times n}$ is the system noise distribution (shaping) matrix. The Kalman filtering process typically involves two distinct phases: the prediction and update steps.

In the prediction step, the *a priori* error state, $\delta\boldsymbol{x}^-$, is considered equal to zero to facilitate linearization, a crucial aspect of the EKF method. The state covariance is then propagated using the known model:

$$\delta\boldsymbol{x}^- = 0 \tag{3}$$

$$\mathbf{P}_k^- = \boldsymbol{\Phi}_{k-1}\mathbf{P}_{k-1}^+\boldsymbol{\Phi}_{k-1}^T + \mathbf{Q}_{k-1} \tag{4}$$

where $\mathbf{P}_k^-$ is the *a priori* state covariance estimate at time $k$ and $\mathbf{P}_{k-1}^+$ is the *a posteriori* state covariance estimate at time $k-1$. The transition matrix, denoted as $\boldsymbol{\Phi}_{k-1}$, is typically derived through a power-series expansion of the system matrix, $\mathbf{F}$, and the propagation interval, $\tau_s$:

$$\boldsymbol{\Phi}_{k-1} = \sum_{r=0}^{\infty}\frac{\mathbf{F}_{k-1}^r}{r!}\tau_s^r \tag{5}$$

and $\mathbf{F}^{k-1}$ is defined as:

$$\mathbf{F}^{k-1} = \frac{\partial \boldsymbol{f}(\boldsymbol{x})}{\partial \boldsymbol{x}}|_{x=x^e} \tag{6}$$

where $\boldsymbol{f}(\boldsymbol{x})$ is the nonlinear function of the state propagation. The exact form of the system noise covariance matrix is (Groves, 2015):

$$\mathbf{Q}_{k-1} = \mathbb{E}\left(\int_{t-\tau_s}^{t}\int_{t-\tau_s}^{t}\boldsymbol{\Phi}_{k-1}(t-t')\mathbf{G}_{k-1}\boldsymbol{n}(t')\,\boldsymbol{n}^T(t'')\mathbf{G}_{k-1}^T\boldsymbol{\Phi}_{k-1}^T(t-t'')\,\partial t'\,\partial t''\right) \tag{7}$$

which is usually approximated and can be written as follows (Gelb et al., 1974):

$$\mathbf{Q}_{k-1} = \frac{1}{2}\left(\boldsymbol{\Phi}_{k-1}\mathbf{G}_{k-1}\mathbf{Q}\mathbf{G}_{k-1}^T + \mathbf{G}_{k-1}\mathbf{Q}\mathbf{G}_{k-1}^T\boldsymbol{\Phi}_{k-1}^T\right)\Delta t. \tag{8}$$

The next phase is the update step, which is executed by the following equations:

$$\mathbf{K}_k = \mathbf{P}_k^-\mathbf{H}_k^T\left(\mathbf{H}_k\mathbf{P}_k^-\mathbf{H}_k^T + \mathbf{R}_k\right)^{-1} \tag{9}$$

$$\mathbf{P}_k^+ = [\mathbf{I} - \mathbf{K}_k\mathbf{H}_k]\mathbf{P}_k^- \tag{10}$$

$$\delta\boldsymbol{x}_k^+ = \mathbf{K}_k\delta\boldsymbol{z}_k \tag{11}$$

where $\mathbf{K}_k$ is the Kalman gain, dictating the balance between incorporating new measurements and the predictions generated by the system's dynamic model. The $\mathbf{H}_k$ and $\mathbf{R}_k$ matrices correspond to the measurement matrix and the covariance matrix of measurement noise, respectively. Lastly, $\delta\boldsymbol{x}_k^+$ and $\mathbf{P}_k^+$ are the *a posteriori* error state and estimated covariance state, respectively. The measurement innovation, $\delta\boldsymbol{z}_k$, is the residual between the state given by the model and the measurement provided by the observation. The measurement matrix, $\mathbf{H}_k$, is defined by:

$$\mathbf{H}_k = \frac{\partial \boldsymbol{h}(\boldsymbol{x})}{\partial \boldsymbol{x}}|_{x=x^e} \tag{12}$$

where $\boldsymbol{h}(\boldsymbol{x})$ represents the update equations characterized by nonlinear functions of the state vector.

### 3.2. Adaptive error-state extended Kalman filter

The conventional ES-EKF initializes the noise covariance matrices, $\mathbf{Q}$ and $\mathbf{R}$, based on prior assumptions about model and measurement uncertainties. Once initialized, these matrices remain constant throughout the estimation process. However, it has been shown in the literature that when the measurement noise covariances are unknown or varying in time, the adaptive approach outperforms the conventional one (Mohamed and Schwarz, 1999). Specifically, we look at the innovation-based adaptive estimation (IAE) approach window, commonly used in different applications (Mehra, 1972). The covariance matrix $\mathbf{Q}_k$ is adjusted as measurements evolve over time. This adaptation of the filter's statistical information matrix is guided by the whiteness of the filter's innovation sequence and takes the following form:

$$\hat{\mathbf{Q}}_k = \mathbf{K}_k\hat{\mathbf{C}}_{v_k}\mathbf{K}_k^T. \tag{13}$$

To calculate $\hat{\mathbf{C}}_{v_k}$, we perform averaging within a moving estimation window of size $N$. This process is carried out as follows:

$$\hat{\mathbf{C}}_{v_k} = \frac{1}{N}\sum_{j=j_0}^{k}\delta\boldsymbol{z}_j\delta\boldsymbol{z}_j^T \tag{14}$$

where $j_0 = k-N+1$ is the first iteration of the process noise covariance estimation. Eq. (13) is used to replace the constant process evaluation (8) inside the conventional ES-EKF.

The approach outlined above represents the fundamental adaptive method. Over time, the literature has introduced various alternative approaches. In this context, we elucidate two of these variations to provide a comparative foundation for the suggested approach. The first approach is described in Ding et al. (2007), Almagbile et al. (2010) and seeks to scale to process noise covariance. The scaling factor is defined as:

$$\beta = \frac{\mathrm{Tr}\{\mathbf{H}_k\left(\boldsymbol{\Phi}_{k-1}\mathbf{P}_{k-1}^+\boldsymbol{\Phi}_{k-1}^T + \hat{\mathbf{Q}}_{k-1}\right)\mathbf{H}_k^T\}}{\mathrm{Tr}\{\mathbf{H}_k\left(\boldsymbol{\Phi}_{k-1}\mathbf{P}_{k-1}^+\boldsymbol{\Phi}_{k-1}^T + \mathbf{Q}_{k-1}\right)\mathbf{H}_k^T\}} \tag{15}$$

and the adaptation rule is defined as:

$$\hat{\mathbf{Q}}_k = \hat{\mathbf{Q}}_{k-1}\sqrt{\beta}. \tag{16}$$

The third adaptive approach utilizes a forgetting factor $\gamma$ to average estimates of $\mathbf{Q}_k$ over time (Akhlaghi et al., 2017):

$$\hat{\mathbf{Q}}_k = \gamma\mathbf{Q}_{k-1} + (1-\gamma)\hat{\mathbf{Q}}_{k-1} \tag{17}$$

where the manually tuned factor $\gamma$ satisfies $0 \leq \gamma \leq 1$. A larger value of $\gamma$ places more weight on previous estimates, incurring less fluctuation in the current estimate but resulting in longer time delays to catch up with changes.

It is worth mentioning that this paper focuses only on the adaptive process noise covariance matrix, as the literature allows for the flexibility to adapt the process noise, measurement noise, or both. Since the Kalman gain — responsible for optimally balancing the uncertainty between the predicted state and the measurement — is proportional to the process noise covariance multiplied by the inverse of the measurement noise covariance, adapting the process noise covariance is sufficient to yield superior results (Bar-Shalom et al., 2001).

## 4. Set-transformer

This section reviews the building blocks of the set-transformer designed to handle time-series data. The adoption of set-transformer for estimating the process noise covariance leverages their intrinsic permutation invariance and robust handling of variable-sized input sets, aligning with the diverse and unordered nature of sensor data inherent in such tasks. This architecture, distinguished by its attention mechanism, adeptly captures complex interdependencies among sensor inputs, facilitating a nuanced understanding of the underlying process dynamics. Consequently, set-transformers offer a sophisticated, end-to-end trainable framework that enhances estimation accuracy and robustness, rendering them highly suitable for advanced navigation systems where precision and adaptability are required.

### 4.1. Patch embedding

Following the approach recommended by Dosovitskiy et al. (2020), the data undergoes a preprocessing step wherein it is partitioned into patches, flattened, and subsequently subjected to a one-dimensional convolution operation. This convolution operation employs a kernel size denoted as $\alpha$ and a stride size as $\beta$, and utilizes patch sizes represented by $\gamma$. The outcome of this patch embedding process is denoted as $\mathbf{x}_p$ and resides in $\mathbb{R}^{N\times D}$, where $N$ signifies the number of samples produced by the one-dimensional convolution layer, and $D$ corresponds to the number of filters employed by the layer and also establishes the latent space dimension that is subsequently input into the set-transformer blocks.

### 4.2. Attention

The attention mechanism is formally defined by the equation

$$Attention(\mathbf{Q},\mathbf{K},\mathbf{V}) = \text{Softmax}\left(\frac{\mathbf{Q}\mathbf{K}^T}{\sqrt{d_q}}\right)\mathbf{V} \in \mathbb{R}^{n\times d_v}. \tag{18}$$

Here, $\mathbf{Q}$ represents $n$ query vectors of size $d_q$ in $\mathbb{R}^{n\times d_q}$, and $\mathbf{K}$ and $\mathbf{V}$ denote $n_v$ key–value pairs in $\mathbb{R}^{n_v\times d_q}$ and $\mathbb{R}^{n_v\times d_v}$, respectively. The dimension of the values vector is denoted as $d_v$.

The Softmax function, as described in Goodfellow et al. (2016), is expressed as

$$\text{Softmax}(Z)i = \frac{e^{Z i}}{\sum_{n=1}^{M} e^{Z_j}}. \tag{19}$$

In this equation, $M$ denotes the number of vectors.

### 4.3. Multihead attention

Multiheads are introduced to extend the capabilities of the attention mechanism. In this model, query vectors ($\mathbf{Q}$), key vectors ($\mathbf{K}$), and value vectors ($\mathbf{V}$) are projected onto $h$ different heads, each having dimensions $\acute{d}_q$, $\acute{d}_q$, and $\acute{d}_v$, respectively. The previously described attention mechanism in (18) is then independently applied to each of these $h$ projections.

This multi-head attention mechanism allows the model to effectively incorporate information from various representation subspaces and positions, enhancing its capacity to capture diverse patterns in the data.

Within this model, trainable parameters $\mathbf{W}_j^Q, \mathbf{W}_j^K \in \mathbb{R}^{d_q\times \acute{d}_q}$, and $\mathbf{W}_j^V \in \mathbb{R}^{d_v\times \acute{d}_v}$ are employed in the following manner:

$$head_j = \text{Attention}(\mathbf{Q}\mathbf{W}_j^Q, \mathbf{K}\mathbf{W}_j^K, \mathbf{V}\mathbf{W}_j^V),\ j = 1,\dots,h. \tag{20}$$

These multiple heads are then linked together in a sequential manner and are scaled by a trainable parameter $\mathbf{W}^O \in \mathbb{R}^{h\acute{d}_v\times d}$, resulting in the following expression:

$$\text{Multihead}(\mathbf{Q},\mathbf{K},\mathbf{V}) = \text{concatenate}(head_1,\dots,head_h)\mathbf{W}^O. \tag{21}$$

A common choice for the hyperparameters in terms of dimensions is $\acute{d}_q = \frac{d_q}{h}$, $\acute{d}_v = \frac{d_v}{h}$, and $d = d_q$. These choices help manage the dimensionality of the multihead attention mechanism effectively.

### 4.4. Set attention block

The multihead attention block (MAB) represents a pivotal modification to the core encoder block of the transformer architecture, as originally introduced by Vaswani et al. (2017). This adaptation, characterized by the removal of positional encoding and dropout operations, is designed to harness the intrinsic power of self-attention mechanisms. In the context of MAB, two matrices, $\mathbf{X}$ and $\mathbf{Y}$, both residing in $\mathbb{R}^{n\times d}$, play a central role. The parameter $\mathbf{L}$ is defined as follows:

$$\mathbf{L} = \text{LayerNorm}(\mathbf{X} + \text{Multihead}(\mathbf{X},\mathbf{Y},\mathbf{Y})). \tag{22}$$

The MAB is an integral component of our architecture, defined as

$$\text{MAB}(\mathbf{X},\mathbf{Y}) = \text{LayerNorm}(\mathbf{L} + \text{FFN}(\mathbf{L})). \tag{23}$$

In this context, FFN denotes a fully connected, feed-forward network. Specifically, it comprises a layer that expands the dimensionality to the size specified by the hyper-parameter $ffe$, which corresponds to the feed-forward expansion. Following the application of a rectified linear unit (ReLU) activation function, the second layer restores the data to its original dimension:

$$\text{FFN}(\mathbf{X}) = \max(0, \mathbf{X}W_1 + b_1)W_2 + b_2 \tag{24}$$

where $W_j$ and $b_j$ are the weights and biases, respectively, such that $j \in \{1,2\}$.

The set attention block (SAB) assumes the crucial role of performing self-attention operations among elements within the set. Importantly, it can be succinctly defined by invoking the MAB:

$$SAB(\mathbf{X}) := \text{MAB}(\mathbf{X},\mathbf{X}). \tag{25}$$

This architectural adaptation facilitates a holistic understanding of relationships among set elements, significantly enhancing the model's representational capacity.

### 4.5. Pooling by multihead attention

This block introduces a learnable aggregation mechanism by establishing a set of $k$ vectors, each with a size of $n$, which are subsequently organized into a matrix denoted as $\mathbf{S} \in \mathbb{R}^{k\times n}$. In this context, assuming that an encoder generates a set of features represented as $\mathbf{Z} \in \mathbb{R}^{n\times d}$, the pooling by multi-head attention block (PMA) is defined as follows:

$$PMA_k(\mathbf{Z}) = MAB(\mathbf{S}, \text{FFN}(\mathbf{Z})) \tag{26}$$

where $k$ represents a tunable hyper-parameter integral to the functionality of this block.

### 4.6. Overall architecture

The architectural framework comprises an encoder followed by a decoder. Given an input matrix $\mathbf{X} \in \mathbb{R}^{n\times d}$, the encoder generates the feature matrix $\mathbf{Z} \in \mathbb{R}^{n\times d}$. The encoder's composition involves the stacking of set attention block (SAB) modules, as indicated in (25):

$$\mathbf{Z} = \text{Encoder}(\mathbf{X}) = SAB_1 \circ \dots \circ SAB_b(\mathbf{X}). \tag{27}$$

Subsequently, the decoder utilizes a learnable aggregation scheme to transform $\mathbf{Z}$ into a set of vectors that traverse a feed-forward network to produce the final outputs:

$$\text{Decoder}(\mathbf{Z}) = \circ \dots \circ \text{FFN}_b(\text{SAB}(PMA_k(\mathbf{Z}))) \in \mathbb{R}^{k\times d} \tag{28}$$

where $b$ signifies the number of stacked blocks. A more detailed overview of the connections between the blocks is presented in the set-transformer original paper. (Lee et al., 2019).

In summary, the critical hyperparameters that require specification within these blocks encompass:

- $\alpha$ - Kernel size for patch embedding
- $\beta$ - Stride size for patch embedding
- $\gamma$ - Patch size for patch embedding
- $D$ - Number of filters in the 1D convolution layer
- $d$ - Latent space dimension, equating to $D$
- $h$ - Number of attention heads
- $ffe$ - Feed-forward expansion
- $b$ - Number of stacked SAB blocks
- $k$ - Number of trainable vectors for aggregation

It is noteworthy that the original set-transformer, initially designed for natural language processing to handle vectorized words within an injective vocabulary, has been adapted for one-dimensional data characterized by infinite values. This adaptation incorporates the patch embedding technique originally proposed for vision transformers. Notably, in this context, a 1D-CNN is employed in lieu of a 2D-CNN.

## 5. Proposed approach

To cope with real-time adaptive process noise covariance matrix estimation, we propose A-KIT, an adaptive Kalman-informed transformer. To this end, we derive a tailored set-transformer network for time series data dedicated to real-time regression of the EKF's process noise covariance matrix. Additionally, a Kalman-informed loss is designed to emulate the principles of the KF, enhancing the accuracy of the process noise covariance. In this manner, A-KIT is designed as a hybrid algorithm combining the strengths of the well-established theory behind EKF and leveraging well-known deep-learning characteristics.

The A-KIT cycle begins with the initialization of the EKF, followed by the execution of the prediction phase in a loop until a valid observation is obtained. Throughout the prediction step, the estimated covariance $\mathbf{P}_k$ is stored in memory, along with all the transition matrices $\mathbf{\Phi}_k$, up until the next update step. Upon encountering a valid observation, the update step is initiated, calculating the innovation-based process noise covariance. Simultaneously, the A-KIT network receives the inertial measurement and estimated state $\boldsymbol{x}^n_{EKF}$ from the filter. It then computes the scaling factor necessary to multiply the innovation-based process noise covariance

$$\mathbf{Q}_k = \hat{\mathbf{Q}}_k \cdot \mathbf{Q}_k^{A-KIT} \tag{29}$$

where $\mathbf{Q}_k^{A-KIT}$ is a diagonal scale factor matrix produced by A-KIT. The Kalman-informed loss, described in Section 5.1.1, utilizes the pre-saved estimated covariance matrix and transition matrices to perform the prediction step once again with the output from the A-KIT. Subsequently, it calculates the Kalman gain to correct the error state. The entire A-KIT training flowchart is depicted in Fig. 1. As A-KIT is intended to regress the process noise, both prediction and update stages of the EKF are present in the cycle, allowing the propagation of the error-state and its associated error-state covariance.

### 5.1. A-KIT algorithm

#### 5.1.1. Kalman-Informed loss

To align the A-KIT network with the model-based EKF, we propose a Kalman-informed loss. To that end, besides the network output, the loss incorporates several EKF parameters. The first is the estimated *a posteriori* state covariance matrix $\mathbf{P}^+_{k-1}$, obtained immediately after a Kalman update. The second set of parameters includes the transition matrices $\mathbf{\Phi}_k$ between updates for model propagation, the measurement matrix $\mathbf{H}_k$ at the time of the update step, the *a priori* state estimate, the measurement innovation $\delta z_k$, the covariance matrix of measurement noise $\mathbf{R}_k$, and the labels, which represent the ground-truth (GT) state.

The loss mirrors the stages of the KF, with the process noise covariance serving as the output of the network, optimized through back-propagation. In essence, the loss function addresses the question: given that sensor fusion is a nonlinear problem, what is the optimal process noise covariance matrix that minimizes the mean squared error of the measured state? The choice to focus on the measured state rather than other navigation parameters stems from its status as an observable state. The algorithm for the Kalman-informed loss is outlined in Algorithm 1.

**Algorithm 1:** Kalman-informed loss

**Input:** $\mathbf{Q}^{A-KIT}$, $\mathbf{P}^+_{k-1}$, $\mathbf{\Phi}_k$, $\mathbf{H}_k$, $\boldsymbol{x}^n_{EKF}$, $\delta \boldsymbol{z}_k$, $\mathbf{R}_k$, $\boldsymbol{x}^n_{GT}$

1 **for** $k \leftarrow 1$ **to** $T$ **do**
2 $\quad \mathbf{P}^-_k = \mathbf{\Phi}_k \mathbf{P}^+_{k-1} \mathbf{\Phi}^T_k + \mathbf{Q}^{A-KIT}$;
3 $\mathbf{K}_T = \mathbf{P}^-_T \mathbf{H}^T_T \left( \mathbf{H}_T \mathbf{P}^-_T \mathbf{H}^T_T + \mathbf{R}_T \right)^{-1}$;
4 $\delta \boldsymbol{x}^+_T = \mathbf{K}_T \delta \boldsymbol{z}_T$;
5 $\boldsymbol{x}^n_{A-KIT} = \boldsymbol{x}^n_{EKF} - \delta \boldsymbol{x}^+_T$;
6 $Loss = MSE\{\boldsymbol{x}^n_{A-KIT}, \boldsymbol{x}^n_{GT}\}$;

**Output:** $Loss$

#### 5.1.2. A-KIT algorithm

Once the network is trained through the aforementioned stages, it can be seamlessly integrated into the EKF methodology before the prediction step to obtain an accurate process noise covariance matrix. The algorithm is detailed in Algorithm 2. Here, $\boldsymbol{x}^n$ is the estimated state, $\delta \boldsymbol{x}^n_s$ is a subset of the error state — which could be the position or velocity pending on the external measurement type — and $\boldsymbol{z}^b$ is the measurement update. A-KIT provides an estimate for the process noise contrivance, enabling the propagation of both error-state and its associated error-state covariance.

**Algorithm 2:** A-KIT over EKF

**Input:** $\hat{\boldsymbol{f}}^b_{ib}$, $\hat{\boldsymbol{\omega}}^b_{ib}$, $\boldsymbol{z}^b$, Innovation window size, $\mathbf{Q}$, $\mathbf{R}$

1 Initialize: $\delta \boldsymbol{x}^n_s$, $\epsilon^n$, $\delta \boldsymbol{b}_a$, $\delta \boldsymbol{b}_g$, $\mathbf{P}$
2 **for** $k \leftarrow 1$ **to** $T$ **do**
3 $\quad \hat{\boldsymbol{f}}^b_{ib} \leftarrow \hat{\boldsymbol{f}}^b_{ib} - \delta \boldsymbol{b}_a$;
4 $\quad \hat{\boldsymbol{\omega}}^b_{ib} \leftarrow \hat{\boldsymbol{\omega}}^b_{ib} - \delta \boldsymbol{b}_g$;
5 $\quad$ Calculate $\mathbf{F}$ (6);
6 $\quad$ Obtain $\mathbf{\Phi}_{k-1}$ (5);
7 $\quad$ Calculate $\mathbf{G}$ (2);
8 $\quad$ Perform the prediction step (4);
9 $\quad$ Solve INS ;
10 $\quad$ **if** *update is valid* **then**
11 $\quad\quad$ Get $\delta \boldsymbol{z}$ ;
12 $\quad\quad$ Calculate $\mathbf{H}_k$ (12);
13 $\quad\quad$ Perform the update step (9)-(11);
14 $\quad\quad$ Correct the INS measurements ;
15 $\quad\quad$ **if** *Num. of updates* $\geq$ *Innovation window size* **then**
16 $\quad\quad\quad \hat{\mathbf{Q}}_k \leftarrow$ innovation covariance (13)-(14) ;
17 $\quad\quad\quad \mathbf{Q}^{A-KIT}_k \leftarrow$ A-KIT($\hat{\boldsymbol{f}}^b_{ib}$, $\hat{\boldsymbol{\omega}}^b_{ib}$,$\hat{\boldsymbol{x}}^n_{EKF}$,$\hat{\mathbf{Q}}_k$);
18 $\quad\quad\quad \mathbf{Q}_k \leftarrow \hat{\mathbf{Q}}_k \cdot \mathbf{Q}^{A-KIT}_k$ (29);

**Output:** $\boldsymbol{x}^n$

## 6. Case study: INS/DVL fusion

Developing a reliable autonomous navigation system is essential for navigating deep underwater environments, particularly in areas beyond human reach. Commonly, underwater navigation relies on the fusion between INS and DVL in an EKF framework, with velocity corrections from the DVL to ensure accurate navigation. The subsequent exploration into navigation components unfold in the following sections: 6.1 reference frames, establishing foundational spatial context; 6.2 dead-reckoning, a method vital for a continuous position, velocity, and orientation deduction; 6.3 DVL velocity estimation, highlighting the role of Doppler velocity logs in enhancing accuracy; and 6.4 conventional EKF and 3.2 adaptive EKF, illuminating advanced filtering strategies essential for refining navigation solutions in complex underwater scenarios.

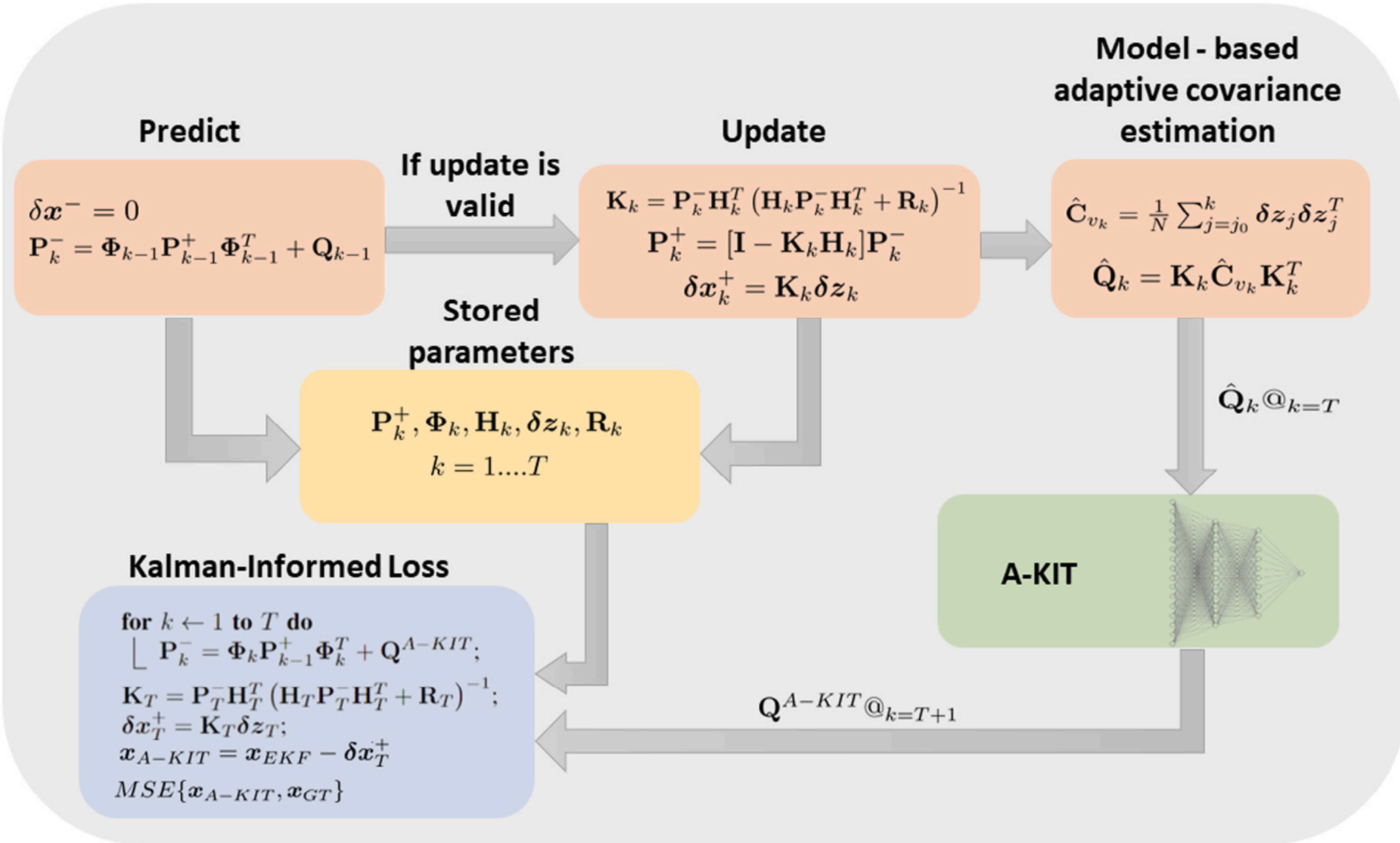

**Fig. 1.** A-KIT training process flowchart illustrating the dynamic estimation of process noise covariance while utilizing the Kalman-informed loss procedure. The parameters shown in the yellow block are stored during the prediction and update steps of the Kalman Filter and are subsequently used in the Kalman-informed loss procedure, along with the scale factor for process noise covariance estimated by A-KIT.

### 6.1. Reference frames

Three reference frames are addressed in this work:

- **Body frame:** The body center is located at the center of mass of the vehicle. The $x$-axis is set along the vehicle's longitudinal axis, pointing forward, the $z$-axis downward, and the $y$-axis pointed outward, completing the right-hand orthogonal coordinate system.
- **Navigation frame:** The navigation or geographic frame is defined locally, relative to the Earth's geoid. We employ the north-east-down (NED) coordinate system where the $x$-axis points to the true north, the $y$-axis points east, and the $z$-axis points toward the interior of the ellipsoid along the ellipsoid normal, the direction of gravity, completing a right-hand orthogonal coordinate system
- **DVL frame:** The DVL frame is a reference frame for the DVL sensor, whose sensitive axes are designed by the manufacturer. The transformation from the DVL frame to the body frame is represented by a known fixed transformation matrix. Rogers (2003).

### 6.2. Dead reckoning

Dead reckoning (DR) is the basic component of underwater navigation. Since GNSS signals do not penetrate water and cannot be used to aid navigation, relying on initial conditions and integrating inertial data over time is necessary. In a terrestrial navigation system, which operates in the local geographic reference frame, the time derivative of the coordinates can be written as

$$\dot{\boldsymbol{r}}^n = \begin{bmatrix} \dot{\varphi} \\ \dot{\lambda} \\ \dot{h} \end{bmatrix} = \begin{bmatrix} \frac{1}{M+h} & 0 & 0 \\ 0 & \frac{1}{(N+h)\cos\varphi} & 0 \\ 0 & 0 & 1 \end{bmatrix} \begin{bmatrix} v_N \\ v_E \\ v_D \end{bmatrix} = \mathbf{D}\boldsymbol{v}^n \tag{30}$$

where $\boldsymbol{r}^n$ represents the position in the geographic frame and $\varphi, \lambda$, and $h$ are the latitude, longitude, and height, respectively, and M and $N$ are radii of curvature in the meridian, and prime vertical (Shin and El-Sheimy, 2002). The velocity dynamics are expressed by

$$\dot{\boldsymbol{v}}^n = \mathbf{C}_b^n \boldsymbol{f}_{ib}^b - (2\boldsymbol{\omega}_{ie}^n + \boldsymbol{\omega}_{en}^n) \times \boldsymbol{v}^n + \boldsymbol{g}^n \tag{31}$$

where $\boldsymbol{v}^n$ is the velocity vector expressed in the navigation frame, $\boldsymbol{f}_{ib}^b$ is the specific force vector expressed in the body frame, $\boldsymbol{g}^n$ is the local gravity vector expressed in the navigation frame, and $\omega_e$ is the magnitude of the rotation rate of the Earth and has the value of $7.29\times10^{-5}\ \frac{rad}{Sec}$. The vectors $\boldsymbol{\omega}_{ie}^n$ and $\boldsymbol{\omega}_{en}^n$ are defined as

$$\boldsymbol{\omega}_{ie}^n = [\omega_e \cos\varphi \quad 0 \quad -\omega_e \sin\varphi]^T \tag{32}$$

$$\boldsymbol{\omega}_{en}^n = [\dot{\lambda}\cos\varphi \quad -\dot{\varphi} \quad -\dot{\lambda}\sin\varphi]^T. \tag{33}$$

The rate of change of the rotation matrix from body to navigation frame is

$$\dot{\mathbf{C}}_b^n = \mathbf{C}_b^n \mathbf{\Omega}_{ib}^b - (\mathbf{\Omega}_{ie}^n + \mathbf{\Omega}_{en}^n)\mathbf{C}_b^n \quad . \tag{34}$$

The body to navigation transformation matrix $\mathbf{C}_b^n$ is defined as

$$\mathbf{C}_b^n = \begin{bmatrix} c\theta c\psi & -c\phi s\psi + s\phi s\theta c\psi & s\phi s\psi + c\phi s\theta c\psi \\ c\theta c\psi & c\phi s\psi + s\phi s\theta c\psi & -s\phi s\psi + c\phi s\theta c\psi \\ -s\theta & s\phi c\theta & c\phi c\theta \end{bmatrix} \tag{35}$$

where sin and cos are denoted as $s$ and $c$, respectively. $\phi, \theta$, and $\psi$ are the three components of the Euler angles: roll, pitch, and yaw, respectively. The notation of $\Omega_a^b$ is a skew-symmetric form of a vector defined as follows:

$$\begin{aligned} \Omega_a^b &= (\omega_a^b \times) = ([\omega_1 \quad \omega_2 \quad \omega_3]\times) = \\ &= \begin{bmatrix} 0 & -\omega_3 & \omega_2 \\ \omega_3 & 0 & -\omega_1 \\ -\omega_2 & \omega_1 & 0 \end{bmatrix} \end{aligned} \tag{36}$$

and operates on $\boldsymbol{\omega}_{ie}^n$ and $\boldsymbol{\omega}_{en}^n$, which are defined in (32)–(33). Additionally, the operator is used on $\boldsymbol{\omega}_{ib}^b$, which is the output of the gyroscope sensor used in (34) (Titterton et al., 2004).

A navigation solution can be established by using Eqs. (30)–(35). Since sensor data is subject to errors such as stochastic noise and bias, the inertial measurements are expressed as follows:

$$\begin{aligned} \hat{\boldsymbol{f}}_{ib}^b &= \boldsymbol{f}_{ib}^b + \boldsymbol{b}_f + \boldsymbol{n}_f \\ \hat{\boldsymbol{\omega}}_{ib}^b &= \boldsymbol{\omega}_{ib}^b + \boldsymbol{b}_\omega + \boldsymbol{n}_\omega \end{aligned} \tag{37}$$

such that the accelerometer and gyro noises are defined as zero-mean white Gaussian noise:

$$\boldsymbol{n}_a \sim \mathcal{N}(0,\sigma_a^2),\ \boldsymbol{n}_g \sim \mathcal{N}(0,\sigma_g^2) \tag{38}$$

and the biases are assumed to be modeled using random walk processes

$$\begin{aligned}\dot{\boldsymbol{b}}_a &= \boldsymbol{n}_{a_b},\ \boldsymbol{n}_{a_b} \sim \mathcal{N}(0,\sigma_{a_b}^2)\\ \dot{\boldsymbol{b}}_g &= \boldsymbol{n}_{g_b},\ \boldsymbol{n}_{g_b} \sim \mathcal{N}(0,\sigma_{g_b}^2)\end{aligned} \tag{39}$$

where $\boldsymbol{b}_a$ and $\boldsymbol{b}_g$ are the biases of the accelerometer and gyroscope, respectively. As the inertial readings (37) are integrated, the navigation solution accumulates error with time. Therefore, the INS is fused with additional sensors (Groves, 2015).

### 6.3. DVL velocity estimation

The DVL is an acoustic-based sensor that transmits four acoustic beams to the seabed in an $\times$ shape configuration, also known as the "Janus Doppler configuration". The sensor acts as both the transmitter and the receiver, and once the acoustic beams are reflected back, due to the Doppler effect, the velocity can be estimated (Brokloff, 1994; Rudolph and Wilson, 2012). In fact, the raw measurements are the velocity in the beam directions and can be expressed in the following manner:

$$\boldsymbol{v}_{beam} = \mathbf{H}_{beam}\boldsymbol{v}^d,\ \mathbf{H}_{beam} = \begin{bmatrix} c\psi_{beam_1}s\alpha & s\psi_{beam_1}s\alpha & c\alpha\\ c\psi_{beam_2}s\alpha & s\psi_{beam_2}s\alpha & c\alpha\\ c\psi_{beam_3}s\alpha & s\psi_{beam_3}s\alpha & c\alpha\\ c\psi_{beam_4}s\alpha & s\psi_{beam_4}s\alpha & c\alpha \end{bmatrix} \tag{40}$$

where sin and cos are denoted as $s$ and $c$, respectively, $\boldsymbol{v}_{beam} \in \mathbb{R}^{4\times1}$ is the velocity in the beam directions, and $\boldsymbol{v}^d \in \mathbb{R}^{3\times1}$ is the velocity in the DVL frame. Each beam transactor is rotated by a yaw angle $\psi_{beam}$ and a pitch angle $\alpha$, such that $\alpha$ is constant and predefined by the manufacturer, and $\psi_{beam}$ is expressed as follows:

$$\psi_{beam_i} = (i-1)\cdot 90^\circ + 45^\circ\ ,\ i = 1,2,3,4. \tag{41}$$

The measurements are not obtained ideally as illustrated in (40) and are subject to built-in errors that are modeled by:

$$\boldsymbol{y} = \mathbf{H}_{beam}[\boldsymbol{v}^d(\mathbf{1} + \boldsymbol{s}_{DVL})] + \boldsymbol{b}_{DVL} + \boldsymbol{n} \tag{42}$$

where $\boldsymbol{b}_{DVL} \in \mathbb{R}^{4\times1}$ is the bias vector, $\boldsymbol{s}_{DVL} \in \mathbb{R}^{3\times1}$ is the scale factor vector, and $\boldsymbol{n} \in \mathbb{R}^{4\times1}$ is a zero mean white Gaussian noise (Tal et al., 2017; Cohen and Klein, 2022).

Once the raw measurements are obtained, the next phase is to extract $\boldsymbol{v}^d$ by filtering the data according to the following cost function:

$$\hat{\boldsymbol{v}}^d = \underset{\boldsymbol{v}^d}{\operatorname{argmin}} \|\ \boldsymbol{y} - \mathbf{H}_{beam}\boldsymbol{v}^d\ \|^2. \tag{43}$$

The solution to (43) is (Bar-Shalom et al., 2001):

$$\hat{\boldsymbol{v}}^d = (\mathbf{H}_{beam}^T\mathbf{H}_{beam})^{-1}\mathbf{H}_{beam}^T\boldsymbol{y}. \tag{44}$$

Finally, the DVL's estimated velocity vector is transformed to the body frame using (Markley, 1988):

$$\hat{\boldsymbol{v}}^b = \mathbf{C}_d^b\hat{\boldsymbol{v}}^d \tag{45}$$

where $\mathbf{C}_d^b$ is the DVL to body transformation matrix and $\hat{\boldsymbol{v}}^b$ is the DVL velocity in the body frame.

### 6.4. INS/DVL fusion

In the case of DVL velocity updates, the position is not directly observable (Klein and Diamant, 2015); hence, a twelve-state vector residual, $\boldsymbol{\delta x^n}$, is employed:

$$\boldsymbol{\delta x^n} = [\boldsymbol{\delta v}^{nT}\quad \boldsymbol{\epsilon}^{nT}\quad \boldsymbol{\delta b}_a^T\quad \boldsymbol{\delta b}_g^T]^T \in \mathbb{R}^{12\times1} \tag{46}$$

where $\boldsymbol{\delta v}^n \in \mathbb{R}^{3\times1}, \boldsymbol{\epsilon} \in \mathbb{R}^{3\times1}, \boldsymbol{\delta b}_a \in \mathbb{R}^{3\times1}$ and $\boldsymbol{\delta b}_g \in \mathbb{R}^{3\times1}$ are the velocity error-states expressed in the navigation frame, the misalignment error, accelerometer bias residual error, and gyroscope bias residual error, respectively. The linearized error state differential equation is

$$\boldsymbol{\delta\dot{x}^n} = \mathbf{F}\boldsymbol{\delta x^n} + \mathbf{G}\boldsymbol{n} \tag{47}$$

where $\boldsymbol{n} \in \mathbb{R}^{12\times1}$ is the system noise vector, $\mathbf{F} \in \mathbb{R}^{12\times12}$ is the system matrix, and $\mathbf{G} \in \mathbb{R}^{12\times12}$ is the system noise distribution matrix. There are several independent sources of noise in the system, each of which is assumed to have a zero mean Gaussian distribution as expressed in (38)–(39) and can be formulated into

$$\boldsymbol{n} = [\boldsymbol{n}_a^T\quad \boldsymbol{n}_g^T\quad \boldsymbol{n}_{a_b}^T\quad \boldsymbol{n}_{g_b}^T]^T \in \mathbb{R}^{12\times1}. \tag{48}$$

The system matrix, $\mathbf{F}$, is given by:

$$\mathbf{F} = \begin{bmatrix} \mathbf{F}_{vv} & \mathbf{F}_{v\epsilon} & \mathbf{0}_{3\times3} & \mathbf{C}_b^n\\ \mathbf{F}_{\epsilon v} & \mathbf{F}_{\epsilon\epsilon} & \mathbf{C}_b^n & \mathbf{0}_{3\times3}\\ \mathbf{0}_{3\times3} & \mathbf{0}_{3\times3} & \mathbf{0}_{3\times3} & \mathbf{0}_{3\times3}\\ \mathbf{0}_{3\times3} & \mathbf{0}_{3\times3} & \mathbf{0}_{3\times3} & \mathbf{0}_{3\times3} \end{bmatrix} \tag{49}$$

and the exact sub-matrices can be found, for example, in Groves (2015). In terms of the system noise distribution, matrix $\mathbf{G}$ is

$$\mathbf{G} = \begin{bmatrix} \mathbf{0}_{3\times3} & \mathbf{C}_b^n & \mathbf{0}_{3\times3} & \mathbf{0}_{3\times3}\\ \mathbf{C}_b^n & \mathbf{0}_{3\times3} & \mathbf{0}_{3\times3} & \mathbf{0}_{3\times3}\\ \mathbf{0}_{3\times3} & \mathbf{0}_{3\times3} & \mathbf{I}_{3\times3} & \mathbf{0}_{3\times3}\\ \mathbf{0}_{3\times3} & \mathbf{0}_{3\times3} & \mathbf{0}_{3\times3} & \mathbf{I}_{3\times3} \end{bmatrix} \tag{50}$$

where $\mathbf{0}_{3\times3}$ is a three-by-three zero matrix and $\mathbf{I}_{3\times3}$ is a three dimensional identity matrix. The measurement innovation, $\boldsymbol{\delta z}_k$, is the residual between the velocity as calculated by the INS velocity (31) and the velocity given by the DVL in the body frame:

$$\boldsymbol{\delta z} = \mathbf{C}_n^b\boldsymbol{v}_{INS}^n - \boldsymbol{v}_{DVL}^b. \tag{51}$$

By deriving the perturbation velocity error (51), the measurement matrix, $\mathbf{H}_{DVL} \in \mathbb{R}^{3\times12}$, can be received:

$$\mathbf{H}_{DVL} = [\mathbf{C}_n^b\quad -\mathbf{C}_n^b(\boldsymbol{v}_{INS}^n\times)\quad \mathbf{0}_{3\times3}\quad \mathbf{0}_{3\times3}]. \tag{52}$$

Once both the prediction and update steps are completed, corrections of the measurements occur in the following manner:

$$\boldsymbol{v}_{EKF}^n = \boldsymbol{v}_{INS}^n - \boldsymbol{\delta v^n} \tag{53a}$$

$$\hat{\mathbf{C}}_b^n = [\mathbf{I} - (\boldsymbol{\epsilon^n}\times)]\mathbf{C}_b^n \tag{53b}$$

$$\hat{\boldsymbol{b}}_a = \boldsymbol{b}_a + \boldsymbol{\delta b}_a \tag{53c}$$

$$\hat{\boldsymbol{b}}_g = \boldsymbol{b}_g + \boldsymbol{\delta b}_g. \tag{53d}$$

### 6.5. Implementation

As mentioned earlier, the proposed architecture, A-KIT, is a set-transformer tailored to process inertial and velocity time series data. The network takes a mini-batch comprising specific force, angular velocity from the IMU, and estimated velocity from the first version of the adaptive EKF. For each, a one-second window was taken, meaning one hundred samples. This data then undergoes the modified patch-embedding layer designed for time series data. Subsequently, it enters the set-transformer, as defined in Section 4. Additionally, the feed-forward network at the end has a slight modification: following the activation function, there is a dropout layer with a probability of 0.1, and the network's output passes through a ReLU activation function, ensuring that output values represent a positive definite matrix. However, any zero values the activation function produces are replaced with the value one. Lastly, the output of A-KIT is multiplied by the intermediate estimated innovation-based covariance matrix provided by (13) to generate the new discrete process noise covariance (29). An illustrative block diagram depicting our A-KIT approach in the INS/DVL case study is visualized in Fig. 2. The hyperparameters used in the modified set-transformer A-KIT are presented in Table 1. The optimizer employed is the RMSprop optimizer with a weight decay of $1 \times 10^{-4}$, a learning rate of $1 \times 10^{-5}$, and a mini-batch size of 400, trained over 100 epochs.

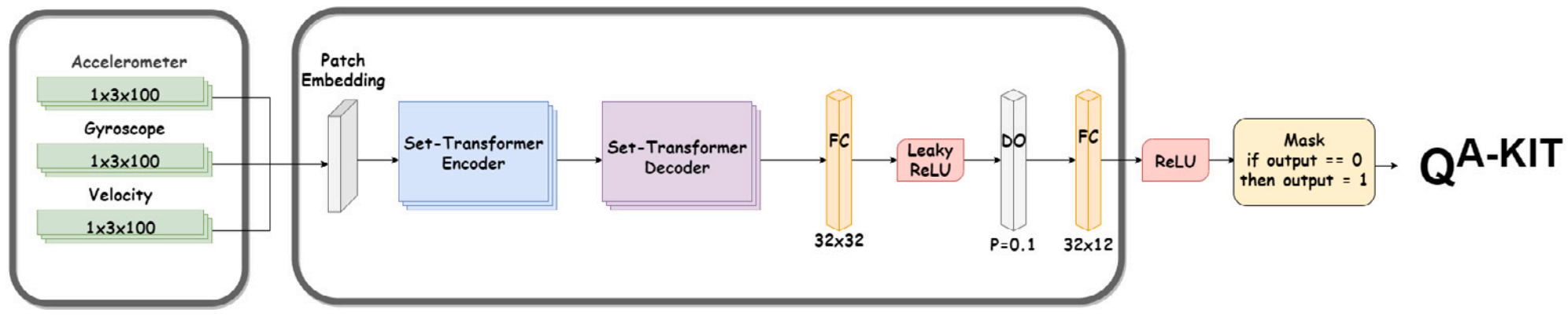

**Fig. 2.** A block diagram illustrating the set-transformer architecture for A-KIT INS/DVL fusion. The network takes a window of 100 samples from the inertial data along with velocity estimations from the EKF as input and produces the diagonal elements of the scale-factor matrix.

**Table 1**
The network's hyper-parameters and values.

| Hyper-parameter | $\alpha$ | $\beta$ | $\gamma$ | $D$ | $h$ | $ffe$ | $b$ | $k$ |
|---|---|---|---|---|---|---|---|---|
| Value | 5 | 1 | 1 | 32 | 2 | 64 | 2 | 1 |

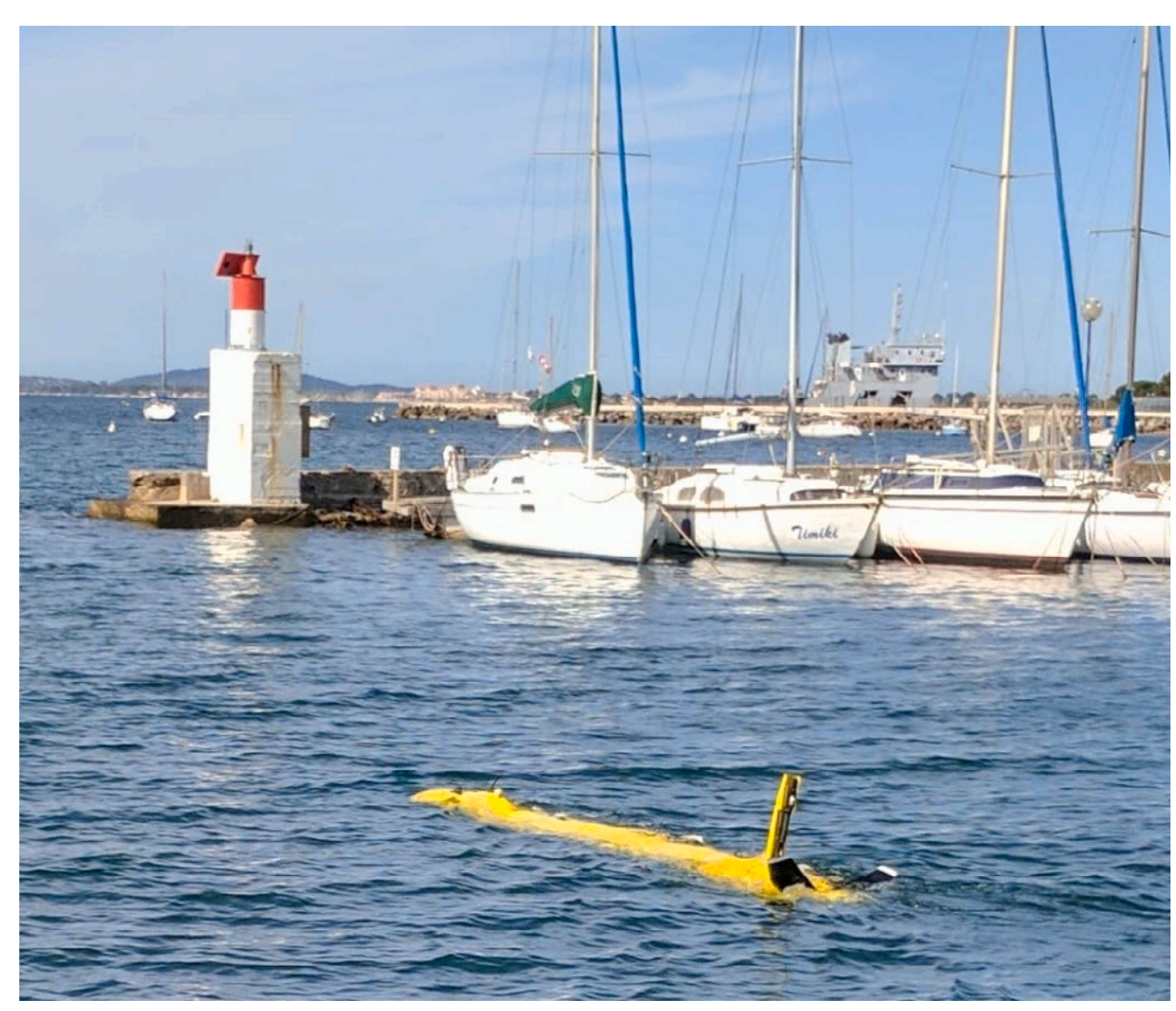

**Fig. 3.** Snapir AUV being dispatched to a mission at sea.

## 7. Analysis and results

### 7.1. Dataset acquisition

We conducted experiments in the Mediterranean Sea near the shore of Haifa, Israel, using the Snapir AUV to gather data. Snapir is a modified ECA Group, A18D mid-size AUV for deep-water applications. It performs autonomous missions up to 3000 meters in depth with 21 h of endurance (ECA Group, 2023). Snapir is equipped with iXblue Phins Subsea, which is a FOG-based high-performance subsea inertial navigation system (iXblue, 2023b) and Teledyne RDI Work Horse navigator DVL (Teledyne Marine, 2023) that achieve accurate velocity measurements with a standard deviation of 0.02 [m/s]. The INS operates at a rate of 100 [Hz] and the DVL at 1 [Hz]. Fig. 3 displays the Snapir AUV during a mission.

The dataset was recorded on June 8th, 2022, and contains approximately seven hours of data with different maneuvers, depths, and speeds. The train set is composed of eleven different data sections, each of a duration of 400 [s] and with different dynamics for diversity. We used only 73.3 min of the recorded data for training. As ground truth (GT), we used the filter solution given by Delph INS, post-processing software for iXblue's INS-based subsea navigation (iXblue, 2023a). To evaluate the approach, we examined an additional two 400 [s] segments of the data that are not present in the training set, referring to them as the test set. The trajectories in the NED frame for each segment in both the training and test sets are illustrated in Fig. 4.

Additionally, to illustrate the difference in trajectory parameters, a radar chart is presented in Fig. 5. The normalized values of average speed, depth, roll, pitch, and yaw highlight the differences between trajectories, facilitating better generalization across common maneuvers performed by the AUV.

### 7.2. Competitive approaches

The three variations of the model-based adaptive EKF (AEKF) were computed using an innovation window size of five updates. The initial $\delta \boldsymbol{x}_0$ and $\mathbf{P}_0^+$ for the conventional EKF are

$$\delta \boldsymbol{x}_0 = \left[0.1[\tfrac{\text{m}}{\text{s}}]\mathbf{1}_3 \quad 2.5^\circ\mathbf{1}_3 \quad 15[\text{m}g]\mathbf{1}_3 \quad 15[\tfrac{^\circ}{\text{h}}]\mathbf{1}_3\right] \tag{54}$$

$$\mathbf{P}_0^+ = \begin{bmatrix} 0.2[\frac{\text{m}}{\text{s}}]\mathbf{I}_3 & \mathbf{0}_3 & \mathbf{0}_3 & \mathbf{0}_3 \\ \mathbf{0}_3 & 5^\circ\mathbf{I}_3 & \mathbf{0}_3 & \mathbf{0}_3 \\ \mathbf{0}_3 & \mathbf{0}_3 & 30[\text{mg}]\mathbf{I}_3 & \mathbf{0}_3 \\ \mathbf{0}_3 & \mathbf{0}_3 & \mathbf{0}_3 & 30[\frac{^\circ}{\text{h}}]\mathbf{I}_3 \end{bmatrix} \tag{55}$$

while for the adaptive EKF variations, these parameters are:

$$\delta \boldsymbol{x}_0 = \left[0.1[\tfrac{\text{m}}{\text{s}}]\mathbf{1}_3 \quad 0.5^\circ\mathbf{1}_3 \quad 15[\text{mg}]\mathbf{1}_3 \quad 0.5[\tfrac{^\circ}{\text{h}}]\mathbf{1}_3\right]. \tag{56}$$

$$\mathbf{P}_0^+ = \begin{bmatrix} 0.2[\frac{\text{m}}{\text{s}}]\mathbf{I}_3 & \mathbf{0}_3 & \mathbf{0}_3 & \mathbf{0}_3 \\ \mathbf{0}_3 & 1^\circ\mathbf{I}_3 & \mathbf{0}_3 & \mathbf{0}_3 \\ \mathbf{0}_3 & \mathbf{0}_3 & 30[\text{mg}]\mathbf{I}_3 & \mathbf{0}_3 \\ \mathbf{0}_3 & \mathbf{0}_3 & \mathbf{0}_3 & 1[\frac{^\circ}{\text{h}}]\mathbf{I}_3 \end{bmatrix} \tag{57}$$

These values were determined through manual optimization to yield the best results with respect to the VRMSE metric. The first AEKF, denoted as AEKF1, is defined in (13), AEKF2 in (16), and AEKF3 in (17). For the latter, a forgetting factor of $\gamma = 0.15$ was chosen.

### 7.3. Evaluation metrics

To assess the performance of the proposed approach, we employed the root mean square error (RMSE) metric, examining both velocity RMSE (VRMSE) and position RMSE (PRMSE), defined as follows:

$$PRMSE(\boldsymbol{p}_i, \hat{\boldsymbol{p}}_i) = \sqrt{\frac{\sum_{i=1}^{N}(\boldsymbol{p}_i - \hat{\boldsymbol{p}}_i)^2}{N}} \tag{58}$$

where $N$ is the number of samples, $\boldsymbol{p}_i$ denotes the ground truth position vector, and $\hat{\boldsymbol{p}}_i$ represents the predicted position vector.

$$VRMSE(\boldsymbol{v}_i, \hat{\boldsymbol{v}}_i) = \sqrt{\frac{\sum_{i=1}^{N}(\boldsymbol{v}_i - \hat{\boldsymbol{v}}_i)^2}{N}} \tag{59}$$

where $\boldsymbol{v}_i$ denotes the ground truth velocity vector, and $\hat{\boldsymbol{v}}_i$ represents the predicted velocity vector.

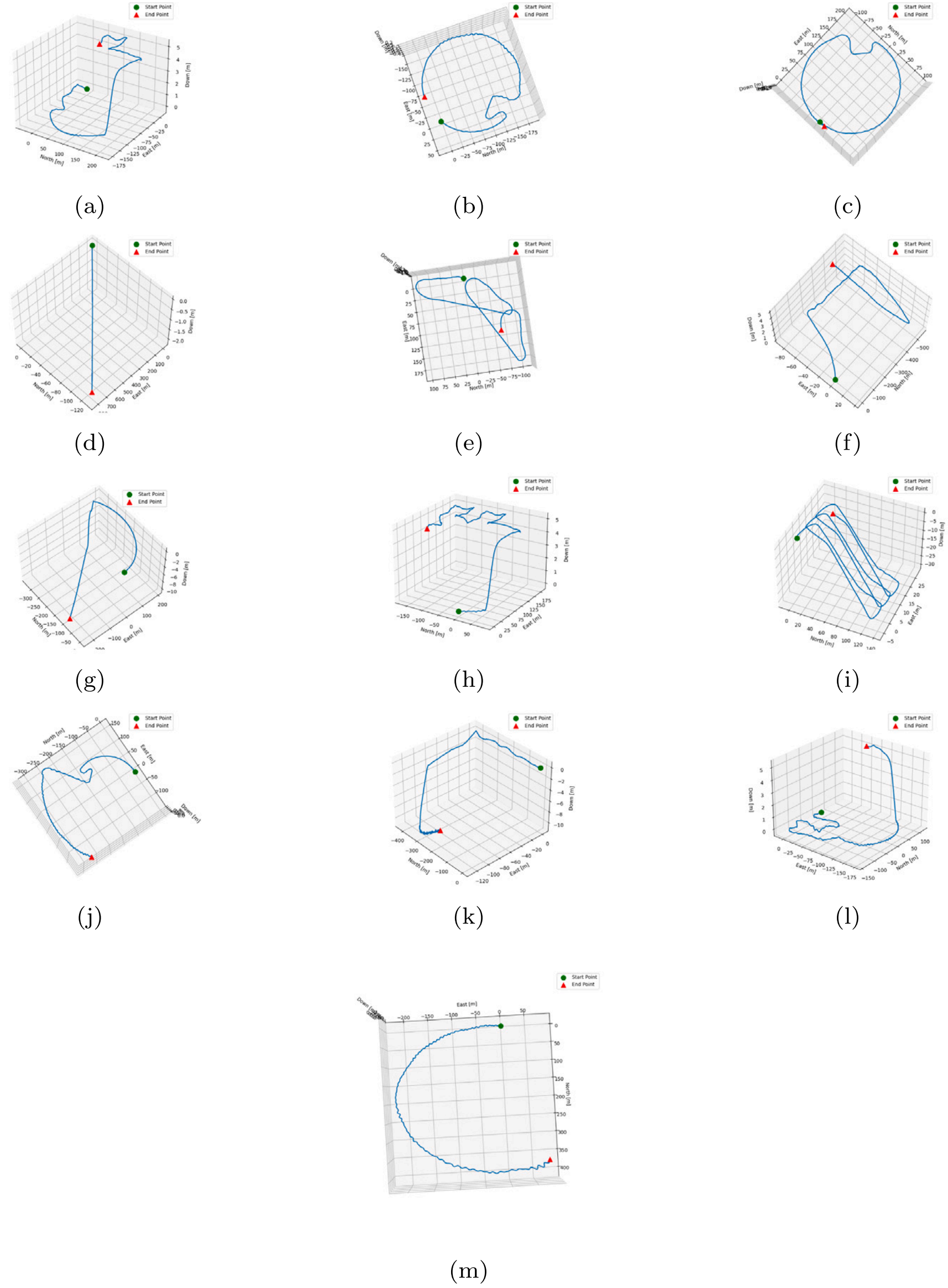

**Fig. 4.** Thirteen AUV trajectories in the NED frame, measured in meters. Green circles denote starting points, and red triangles signify endpoints. Trajectories (a)–(k) belong to the training set, while (l)–(m) comprise the test set.

### 7.4. Experimental results

The proposed A-KIT approach underwent evaluation using test set trajectories depicted in Fig. 4, comparing its performance against the conventional EKF and three AEKFs. A Monte Carlo (MC) test consisting of 100 runs was made, with initial conditions drawn from a Gaussian distribution and $\mathcal{N}(\delta \boldsymbol{x}_0, \mathbf{P}_0)$ as defined in (54)–(57). Figs. 6(b) and 7(b) display the position comparison for each method for test trajectories (l) and (m). In Fig. 6(a), the cumulative mean of the VRMSE for trajectory (l) is presented as a function of the MC iterations. Convergence to a steady state is observed for all approaches, with the A-KIT outperforming both conventional and adaptive EKF versions. The A-KIT demonstrates a 67.8% improvement in VRMSE and a 49.5% improvement in PRMSE compared to the EKF. Additionally, it exhibits an improvement of more than 10% and 8% for the VRMSE and PRMSE of the adaptive approaches, respectively. In the zoomed-in portion of 6(b), the suggested approach is shown to closely follow the trajectory compared to the others. When examining trajectory (m), Fig. 7(a) demonstrates similar behavior as for the (l) trajectory. All the filters converge to a steady state, and the A-KIT possesses the lower cumulative mean VRMSE. In this case, the results are even superior to those of the previous ones. An improvement of more than 90% of the A-KIT over the EKF with respect to the VRMSE and 87.7% with respect to the PRMSE is observed. Furthermore, there is an improvement of more than 31% for the A-KIT when compared to the VRMSE of the adaptive approaches and more than 37% when compared to the PRMSE. All the results are organized in Table 2, where the exact values of the VRMSE and PRMSE are provided.

### 7.5. Real-time computation assessment

Extensive research has explored the real-time implementation of EKF and AEKF, with their computational demands largely dependent on the number of states (Giroux et al., 2005). Studies show that EKF generally requires less computational power than nonlinear filters like the UKF and Particle Filter. The key distinction between the conventional approach and our proposed method lies in the EKF and AEKF update phase. In our framework, after the network is trained, only forward propagation is necessary during the update step. This raises the question of whether the suggested method can fulfill the requirements of a real-time navigation system. Given that INS/DVL EKF systems are known to function in real-time, we benchmarked the execution time of a single prediction and update cycle for the competing approaches and our A-KIT model. This evaluation involved running the cycle 10,000 times, with the average and standard deviation of the computation times recorded. The algorithms were written in Python 3.9.7, with the A-KIT deep learning model implemented using PyTorch 1.10.2. The computational environment consisted of an 11th Gen Intel® Core™ i7-1165G7 CPU running at 2.80 GHz, equipped with 8 cores and 16 GB of RAM.

**Table 2**
PRMSE and VRMSE performance of the conventional EKF (fixed process noise covariance), three AEKFs, and our proposed approach.

| Method | VRMSE [m/s] | | A-KIT VRMSE Improvement | | PRMSE [m] | | A-KIT PRMSE Improvement | |
|---|---|---|---|---|---|---|---|---|
| Trajectory | (l) | (m) | (l) | (m) | (l) | (m) | (l) | (m) |
| EKF | 0.977 | 0.694 | 67.8% | 90.7% | 35.5 | 80.05 | 49.5% | 87.7% |
| AEKF ver. 1 | 0.358 | 0.0934 | 12.29% | 31.47% | 19.6 | 15.567 | 8.6% | 37.11% |
| AEKF ver. 2 | 0.555 | 0.503 | 43.4% | 87.2% | 26.1 | 66.4 | 31.4% | 85.2% |
| AEKF ver. 3 | 0.350 | 0.0931 | 10.28% | 31.2% | 20.7 | 15.54 | 13.5% | 37% |
| **A-KIT (Ours)** | **0.314** | **0.064** | N/A | N/A | **17.9** | **9.79** | N/A | N/A |

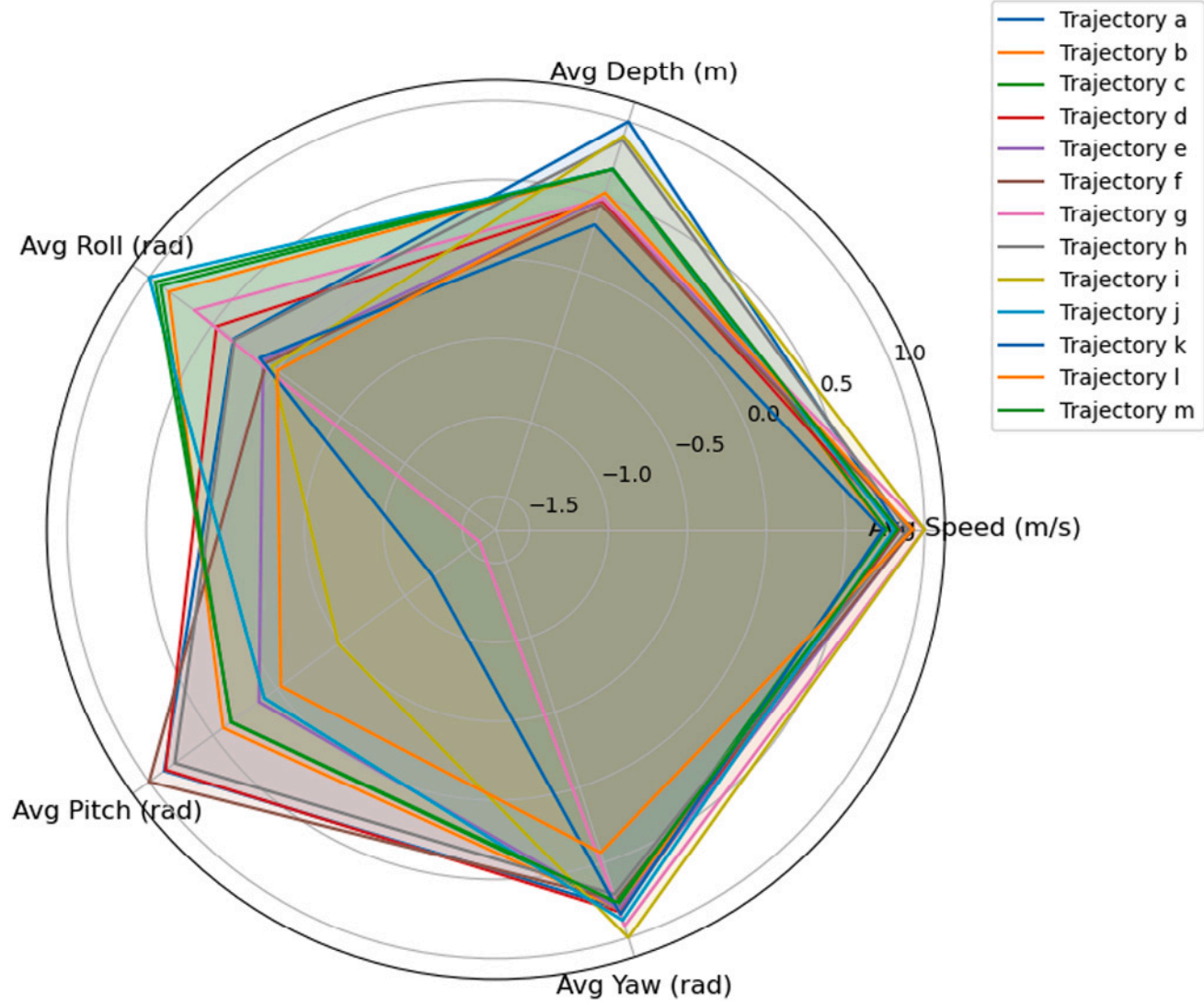

**Fig. 5.** Radar chart illustrating the average values of dynamic characteristics across 13 trajectories. The metrics visualized include average speed (m/s), depth (m), roll (rad), pitch (rad), and yaw (rad). Each trajectory is represented as a distinct polygon, showcasing the diversity in dynamics and ensuring comprehensive representativeness of the training and testing datasets.

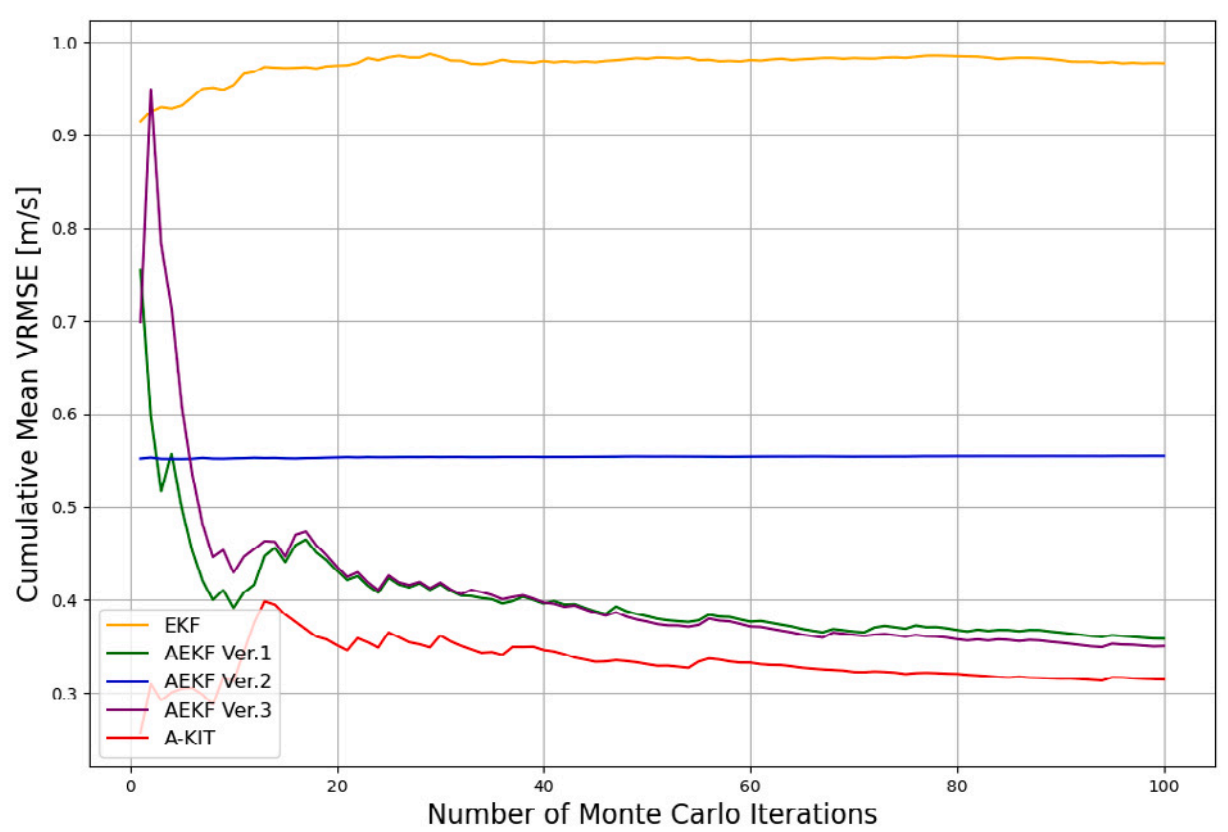

(a)

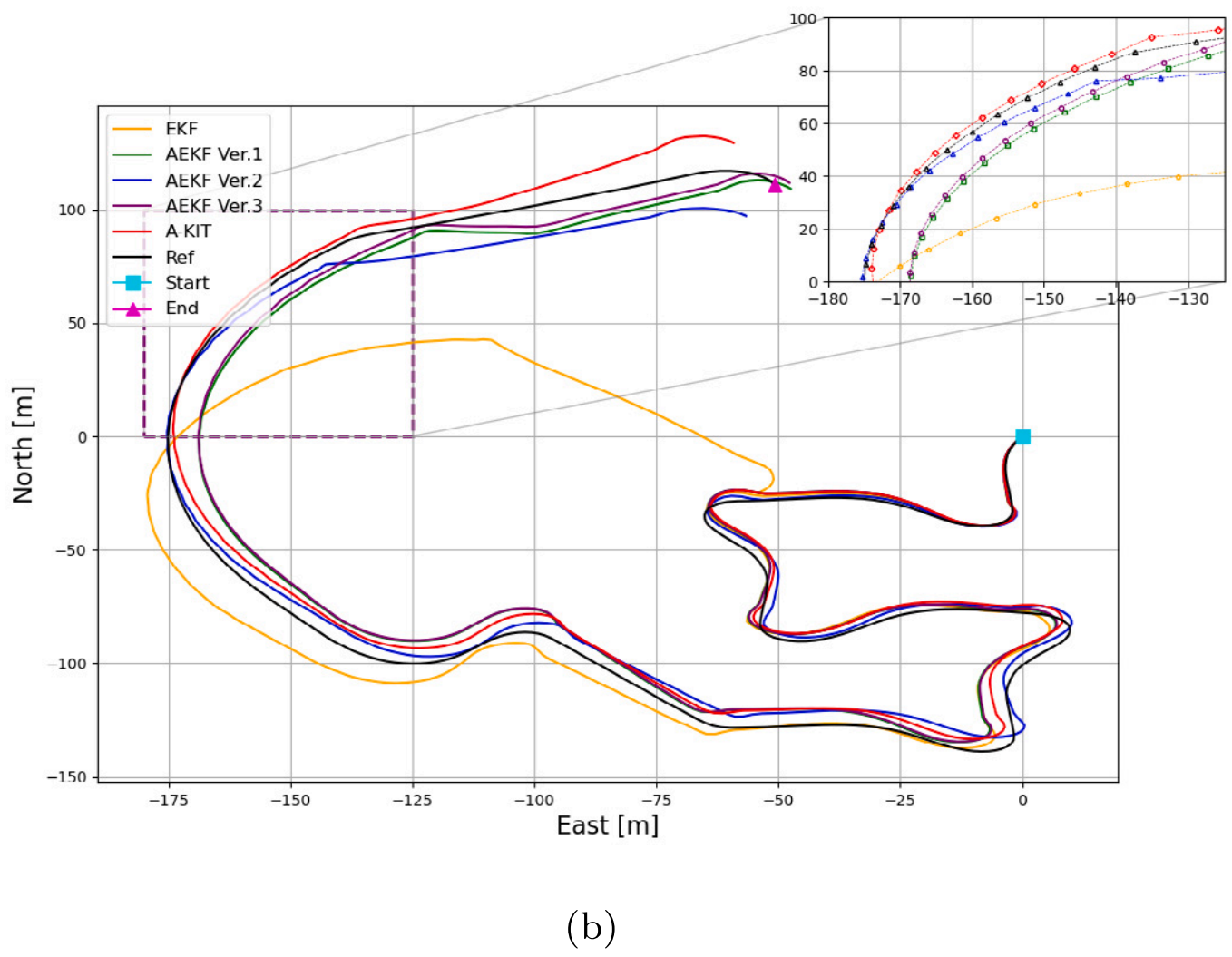

(b)

**Fig. 6.** (a) The cumulative mean of the VRMSE as a function of the Monte Carlo iteration for the (l) trajectory. In (b), a trajectory comparison is presented between the EKF variations for the (l) trajectory.

The runtime results, as presented in Fig. 8, demonstrate that the proposed A-KIT method has a longer runtime of approximately 8,846 μs per cycle compared to the baseline EKF and adaptive EKF variations (AEKF Ver.1, Ver.2, and Ver.3), which range between 813 and 975 μs. Despite this increased runtime, the A-KIT method meets the real-time requirements for INS sampling at 100 Hz, as its runtime is well within the 10 ms (10,000 μs) deadline, leaving sufficient buffer time for additional processes. Furthermore, the method satisfies the computational demands of the aiding DVL, which commonly operates at 1 Hz and requires completion within a 1-second window. The additional computational overhead in A-KIT is justified by its enhanced accuracy and robustness in handling sensor errors and environmental uncertainties, making it highly suitable for real-time navigation, particularly in challenging scenarios. This trade-off between computation time and performance highlights A-KIT's capability to deliver real-time navigation support while significantly improving reliability and adaptability. Regarding the additional space required compared to traditional methods, the increase is solely attributed to storing the pre-trained weights. Upon calculation, we found that using A-KIT demands an additional 652.79 KB of storage.

## 8. Conclusions

This paper addresses the crucial task of applying an online adaptive navigation filter. To that end, we proposed A-KIT, a hybrid learning framework. Our approach leverages the advantages of DL, specifically a set-transformer network, to dynamically adapt the process noise covariance matrix of an EKF. We rely on the well-established EKF theory and enjoy the benefits of merging deep-learning algorithms with the filter.

The proposed Kalman-informed loss enables seamless integration between the DL approach and the model-based EKF. Utilizing this loss, the network can be optimized to provide a covariance noise matrix that minimizes errors in the desired state while adhering to all Kalman methodologies.

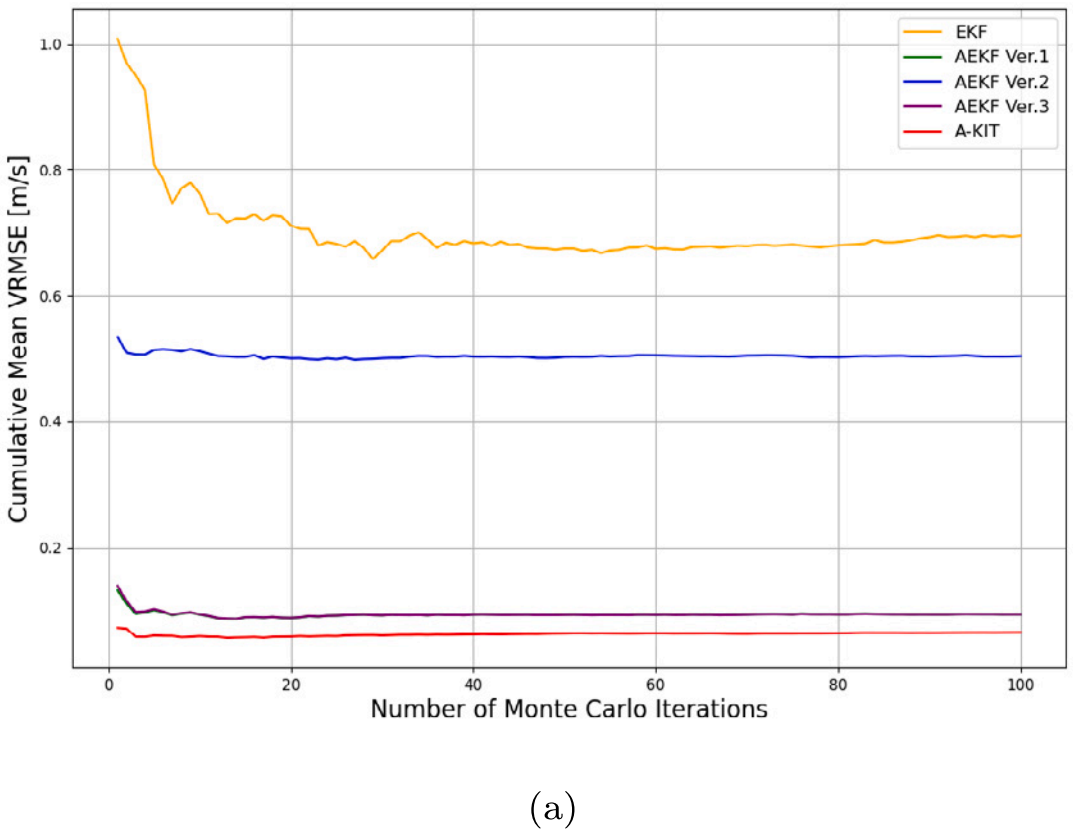

(a)

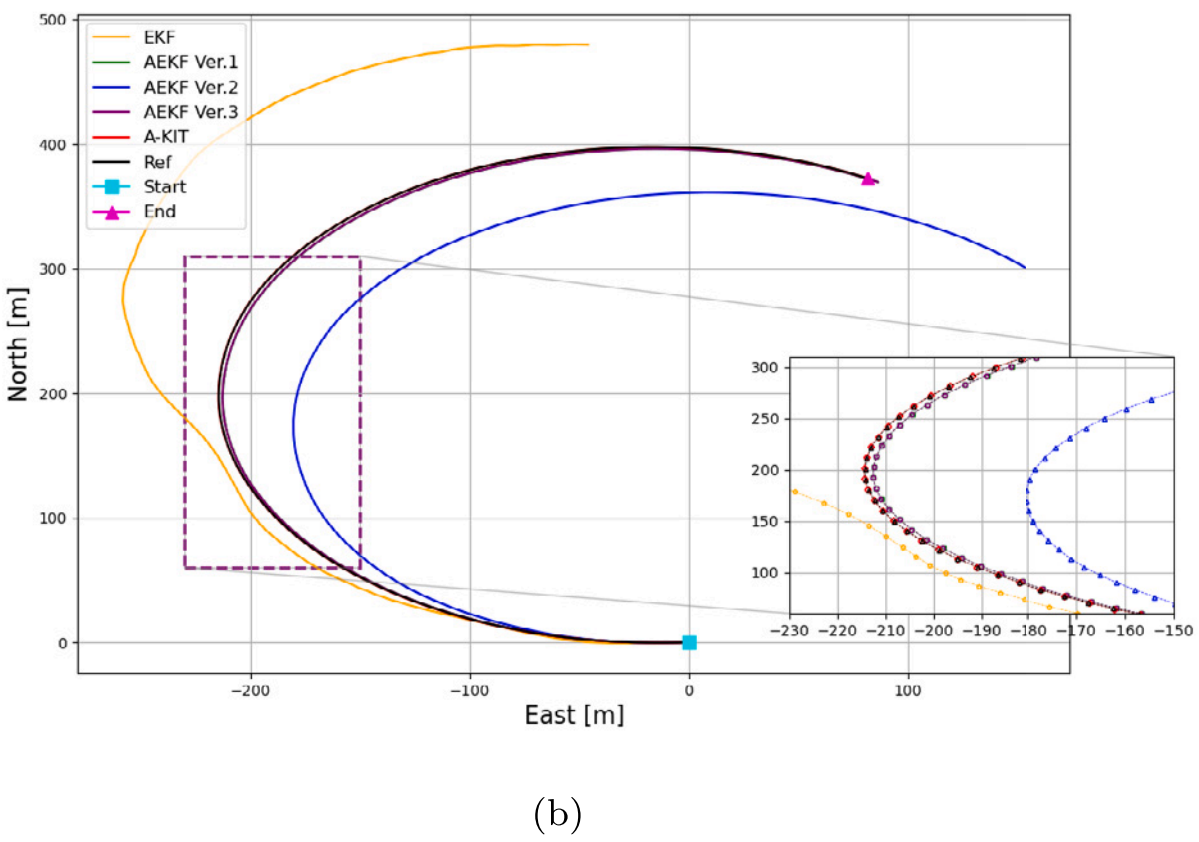

(b)

**Fig. 7.** (a) The cumulative mean of the VRMSE as a function of the Monte Carlo iteration for the (m) trajectory. In (b), a trajectory comparison is presented between the EKF variations for the (m) trajectory.

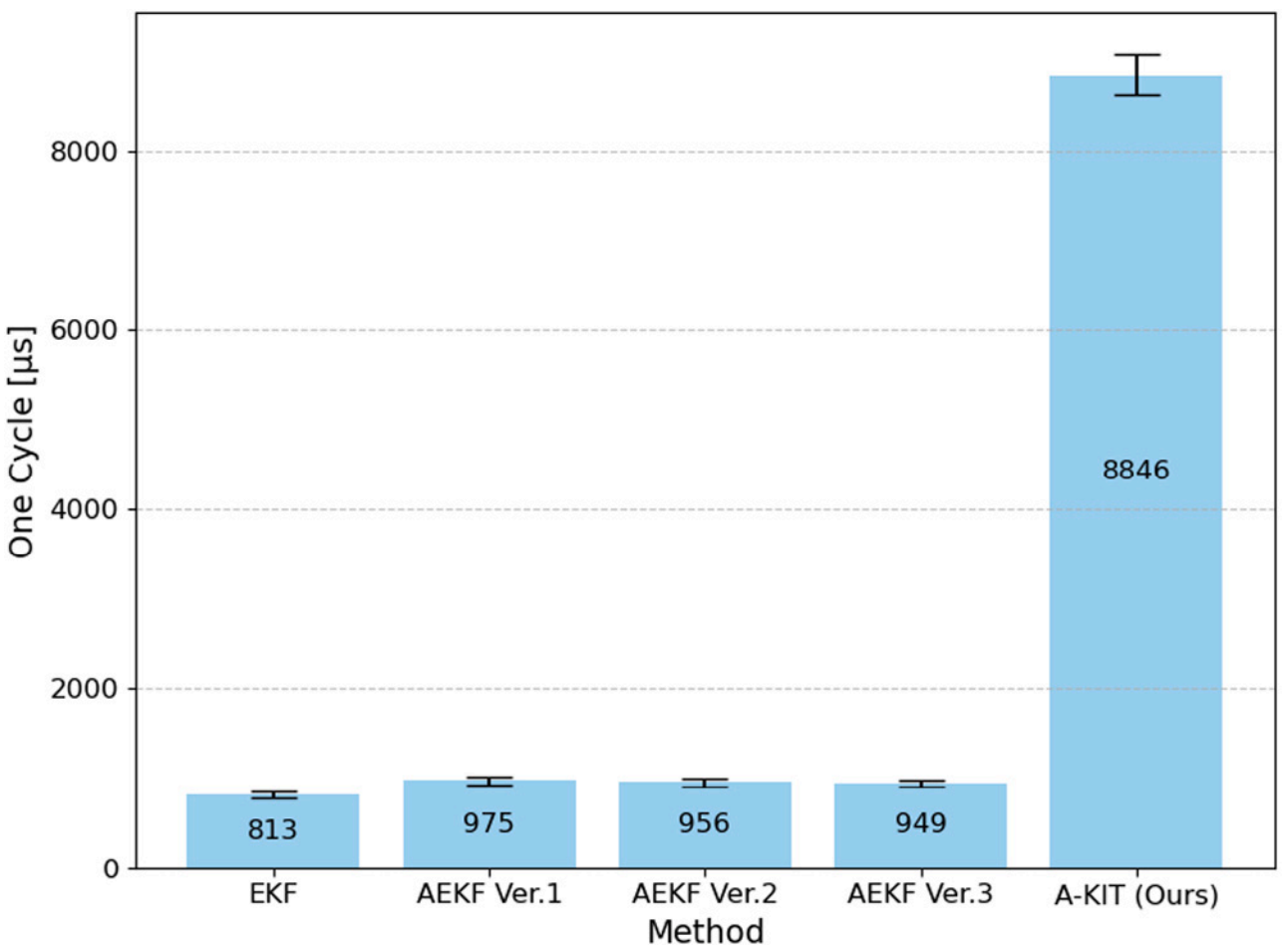

**Fig. 8.** The computation time for a single prediction and update cycle of the INS/DVL fusion process was measured across the compared approaches. Each cycle was executed 10,000 times, and the results are reported as the mean computation time along with its associated standard deviation.

A-KIT was developed as a versatile algorithm applicable to various navigation-related fusion tasks. To demonstrate its effectiveness, we conducted a case study focused on the nonlinear fusion of INS/DVL in an AUV. A-KIT was compared against four approaches, including the conventional EKF with a constant process noise covariance along the trajectory and three different adaptive EKFs that dynamically adjust the process noise covariance based on the filter's innovation. A-KIT was trained and evaluated using real recorded data from an AUV spanning 86.6 min. The results illustrate that A-KIT outperformed all other adaptive approaches, showcasing an average improvement of 35.4% in PRMSE and 35.9% in VRMSE.

This study demonstrates that our A-KIT framework is capable of meeting the demands of real-time navigation tasks. However, computational time analysis revealed that its implementation may require specialized hardware to address the increased processing demands. To build on these findings, future work could explore adapting the method to alternative Kalman filter frameworks, such as the UKF. Additionally, further efforts should focus on improving computational efficiency to reduce execution time while maintaining or enhancing overall performance.

## CRediT authorship contribution statement

**Nadav Cohen:** Writing – review & editing, Writing – original draft, Software, Formal analysis. **Itzik Klein:** Writing – review & editing, Supervision.

## Declaration of competing interest

The authors declare that they have no known competing financial interests or personal relationships that could have appeared to influence the work reported in this paper.

## Acknowledgment

N.C. is supported by the Maurice Hatter Foundation and University of Haifa presidential scholarship for outstanding students on a direct Ph.D. track.

## Data availability

A GitHub repository containing experiment data and code for implementing the A-KIT architecture is available.